\pdfoutput=1

\documentclass[11pt]{article}

\usepackage{natbib} 
\usepackage{fullpage}
\usepackage{amsmath,amsthm,amssymb}
\usepackage{comment}
\usepackage{graphicx}

\usepackage[colorlinks=true, linkcolor=red, urlcolor=blue, citecolor=blue]{hyperref}

\usepackage{times}
\usepackage{latexsym}

\usepackage[T1]{fontenc}
\usepackage[utf8]{inputenc}

\usepackage{microtype}

\usepackage{inconsolata}

\usepackage{graphicx} 
\usepackage{fullpage}

\usepackage{authblk}
\usepackage{url}
\usepackage{tabularx}
\usepackage{siunitx}
\usepackage{booktabs}
\usepackage{xcolor}
\usepackage{comment}
\usepackage{enumitem}
\usepackage{dsfont}
\usepackage{bm}

\usepackage[most]{tcolorbox}
\usepackage{listings}
\definecolor{mygreen}{rgb}{0,0.6,0}
\definecolor{aquamarine}{rgb}{0.5, 1.0, 0.83}
\definecolor{capri}{rgb}{0.0, 0.75, 1.0}
\usepackage[nameinlink,capitalise,noabbrev]{cleveref}

\usepackage{xcolor}
\definecolor{dark-blue}{rgb}{0.15,0.15,0.4}
\hypersetup{
    colorlinks=true,
    linkcolor=blue,
    filecolor=magenta,
    urlcolor=cyan,
    citecolor=dark-blue,
    pdftitle={Smaug: Fixing Failure Modes of Preference Optimisation with DPO-Positive},
    pdfpagemode=FullScreen,
}

\newcommand{\E}{\mathbb{E}}

\newcommand{\model}{Smaug}
\newcommand{\methodShort}{DPOP}

\title{Smaug: Fixing Failure Modes of Preference Optimisation with DPO-Positive}

\author{Arka Pal\thanks{Correspondence to: \texttt{arka.pal@gmail.com}}, Deep Karkhanis, Samuel Dooley, Manley Roberts, Siddartha Naidu, Colin White
    \vspace*{2mm} \\
    Abacus.AI
}

\date{}

\begin{document}

\maketitle

\begin{abstract}
Direct Preference Optimisation (DPO) is effective at significantly improving the performance of large language models (LLMs) on downstream tasks such as reasoning, summarisation, and alignment. Using pairs of preferred and dispreferred data, DPO models the \textit{relative} probability of picking one response over another. In this work, first we show theoretically that the standard DPO loss can lead to a \textit{reduction} of the model's likelihood of the preferred examples, as long as the relative probability between the preferred and dispreferred classes increases. We then show empirically that this phenomenon occurs when fine-tuning LLMs on common datasets, especially datasets in which the edit distance between pairs of completions is low. Using these insights, we design DPO-Positive (DPOP), a new loss function and training procedure which avoids this failure mode. Surprisingly, we find that DPOP outperforms DPO and other fine-tuning procedures across a wide variety of datasets and downstream tasks, including datasets with high edit distances between completions. Furthermore, we find that the DPOP-tuned model outperforms the DPO-tuned model (all else equal) on benchmarks independent of the fine-tuning data, such as MT-Bench. Finally, using DPOP, we create and open-source Smaug-34B and Smaug-72B, with the latter becoming the first open-source LLM to surpass an average accuracy of 80\% on the HuggingFace Open LLM Leaderboard.
\end{abstract}

\section{Introduction} \label{sec:intro}

Aligning large language models (LLMs) with preference data is important for their fluency and applicability to many tasks, with the natural language processing literature using many techniques to incorporate either human or ground-truth feedback \citep{christiano2017deep, stiennon2020learning,ouyang2022training}.
Typically in LLM alignment, we first collect large amounts of preference data, consisting of a context and two potential completions; one of these is labelled as the preferred completion, and the other as the dispreferred.
We use this data to learn a general policy for generating completions in a given context.
Direct Preference Optimisation (DPO) \citep{rafailov2023direct} is a popular method for learning from preference data, and it has shown to be effective at  improving the performance of pretrained LLMs on downstream tasks such as reasoning, summarisation, and alignment \citep{wang2023making,tunstall2023zephyr}.
The theoretical motivation for DPO is based on a preference-ranking model with an implicit reward function that models the \textit{relative} probability of picking the preferred completion over the dispreferred.

\begin{figure*}[t]
    \centering
    \includegraphics[width=0.99\textwidth]{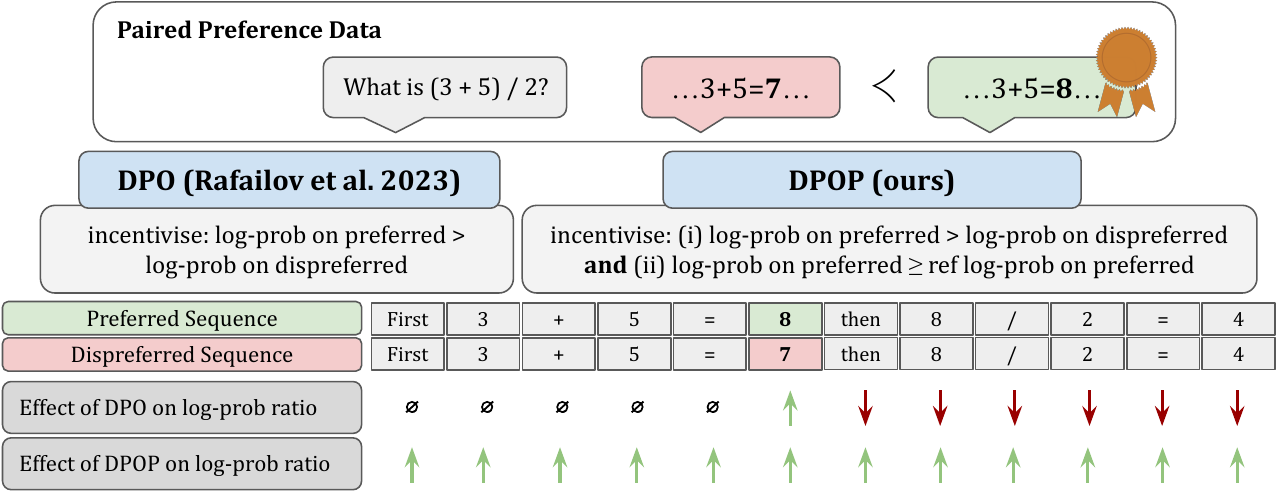}
    \caption{\textbf{\methodShort{} avoids a failure mode of DPO.}
    When preference pairs differ on only a few tokens, DPO receives no loss incentive for the early tokens, and a loss incentive that in some cases can lead to degradation of the log-probs of later tokens (\cref{sec:failure}). We introduce \methodShort, which adds a new term to the loss which leads every token to be incentivised toward the preferred completion (\cref{sec:dpop}).}
    \label{fig:dpop_overview}
\end{figure*}

In this work, first we show theoretically that the standard DPO loss can lead to a \textit{reduction} of the model's likelihood of the preferred completions, as long as the relative probability between the preferred and dispreferred classes increases.
Our theoretical analysis suggests that the problem occurs most frequently in preference datasets with small edit distances between each pair of completions.
Specifically, we show that in the case of preferred and dispreferred examples that differ by few tokens, DPO increases the probability of the token(s) that differ, yet decreases the probability of subsequent tokens (see \cref{fig:dpop_overview} for an overview).
We also present an empirical token-level analysis that matches our theoretical findings on common datasets. 

Using these insights, we design a new loss function: DPO-Positive (\methodShort), which adds a new term to the loss function that penalises reducing the probability of the positive completions.
We also create new preference datasets based on ARC \citep{clark2018think}, HellaSwag \citep{zellers2019hellaswag}, and MetaMath \citep{yu2023metamath} and use them along with DPOP to create new models. 

We introduce the \model{} 
%\footnote{Model name is redacted for anonymity.} 
class of models which use DPOP to achieve state-of-the-art open-source performance.
We fine-tune 7B, 34B, and 72B models on our new datasets and show that \methodShort{} far outperforms DPO. 
We evaluate our resulting models on multiple benchmarks including the HuggingFace Open LLM Leaderboard \citep{open-llm-leaderboard,eval-harness}, which aggregates six popular benchmarks such as MMLU \citep{hendrycks2021measuring} and GSM8K \citep{cobbe2021training}.
On the HuggingFace Open LLM Leaderboard, \model-72B achieves an average accuracy of 80.48\%, becoming the first open-source LLM to surpass an average accuracy of 80\% and improving by nearly 2\% over the second-best open-source model.

In order to address potential concerns about pervasive contamination on the HuggingFace Open LLM Leaderboard, we use an open-source contamination checker, finding that our model scores similarly to popular existing models. Then, we show that \methodShort{} outperforms DPO in an apples-to-apples comparison on an LLM-judged benchmark that is independent of the fine-tuning data: MT-Bench \citep{zheng2023judging}, a challenging benchmark representing eight different categories.
We release our code, models, datasets, and documentation at 
\url{https://github.com/abacusai/smaug}.
%\url{https://anonymous.4open.science/r/dpop}.

\noindent\textbf{Our contributions.} We describe our main contributions below.

\begin{itemize}[topsep=2pt, itemsep=2pt, parsep=0pt, leftmargin=5mm]
  \item We theoretically and empirically show a surprising failure mode of DPO: running DPO on preference datasets with small edit distances between completions can result in a catastrophic decrease in accuracy.  
  \item We introduce DPO-Positive (\methodShort) which we theoretically and empirically show ameliorates the performance degradation. In particular, \methodShort{} often outperforms DPO, even on preference datasets with high edit distances between completions.  
  \item We create new preference-based versions of ARC, HellaSwag, and MetaMath.
  \item Using \methodShort{} and our new datasets, we create the \model{} series of models, with \model{}-72B becoming the first open-source model to achieve an average accuracy of 80\% on the HuggingFace Open LLM Leaderboard. We open-source our trained models, datasets, and code.
\end{itemize}

\section{Background and Related Work}
\label{sec:background}

Large language models (LLMs) have shown impressive zero-shot and few-shot performance \citep{gpt2,gpt3,bubeck2023sparks}.
Recently, researchers have fine-tuned pretrained LLMs on downstream tasks by using human-written completions 
\citep{chung2022scaling,mishra2021cross}
or by using datasets labelled with human-preferred completions relative to other completions \citep{ouyang2022training, bai2022training, ziegler2020finetuning}.
These techniques have been used to improve performance on a variety of downstream tasks such as translation \citep{kreutzer2018reliability} and summarisation \citep{stiennon2020learning}, as well as to create general-purpose models such as Zephyr \citep{tunstall2023zephyr}.
Two of the most popular techniques for learning from preference data are reinforcement learning from human feedback (RLHF) \citep{ouyang2022training, bai2022training, ziegler2020finetuning} and direct preference optimisation (DPO) \citep{rafailov2023direct}.
We summarise these approaches below.

\paragraph{RLHF} Consider a dataset of pairwise-preference ranked data $\mathcal{D} = \{ x^{(i)}, y_w^{(i)}, y_l^{(i)} \}_{i=1}^{N}$ where $x^{(i)}$ are prompts and $y_w^{(i)}$ and $y_l^{(i)}$ are respectively the preferred and dispreferred completions conditioned on that prompt. We have an initial LLM $\pi_{\text{ref}}$ that parameterises a distribution $\pi_{\text{ref}}(y | x)$. Often, we initialise $\pi_{\text{ref}}$ as an LLM that has undergone supervised fine-tuning (SFT) to improve performance on downstream task(s).
RLHF begins by modelling the probability of preferring $y_w$ to $y_l$ using the Bradley-Terry model \citep{BradleyTerry1952} which posits the following probabilistic form:
\begin{equation}
p(y_w \succ y_l | x) = \sigma (r(x, y_w) - r(x, y_l)) \nonumber
\end{equation}
where $\sigma$ is the logistic function and $r(x, y)$ corresponds to some latent reward function that is assumed to exist for the completion $y$ given the prompt $x$.
Given $\mathcal{D}$, we can learn a parameterised estimate of $r$ by minimising the negative log-likelihood of the dataset:
\begin{equation*}
\mathcal{L}_R(r_{\phi}, \mathcal{D}) = -\mathbb{E}_{(x,y_w,y_l) \sim \mathcal{D}} \left[ \log(\sigma(r_{\phi}(x, y_w)
- r_{\phi}(x, y_l))\right].
\end{equation*}
For RLHF, we use reinforcement learning to optimise based on this learned reward function $r_{\phi}$ (with a regularising KL-constraint to prevent model collapse), and obtain a new LLM distribution $\pi_{\theta}$.

\paragraph{DPO} \citet{rafailov2023direct} showed that it is possible to optimise the same KL-constrained reward function as in RLHF without having to learn an explicit reward function. Instead, the problem is cast as a maximum likelihood optimisation of the distribution $\pi_{\theta}$ directly, with the objective:
\begin{equation} \label{eq:dpo}
\mathcal{L}_{\text{DPO}}(\pi_{\theta}; \pi_{\text{ref}}) = -\mathbb{E}_{(x, y_w, y_l) \sim D}
\left[ \log \sigma \left( \beta \log \frac{\pi_{\theta}(y_w | x)}{\pi_{\text{ref}}(y_w | x)} - \beta \log \frac{\pi_{\theta}(y_l | x)}{\pi_{\text{ref}}(y_l | x)} \right) \right]
\end{equation}
where $\beta$ is a regularisation term corresponding to the strength of KL-regularisation in RLHF. In this case, the implicit reward parameterisation is $r(x, y) = \beta \log \frac{\pi_{\theta}(y|x)}{\pi_{\text{ref}}(y|x)}$, and \citet{rafailov2023direct} further showed that all reward classes under the Plackett-Luce model \citep{plackett1975analysis,luce2005individual} (such as Bradley-Terry) are representable under this parameterisation.
For an abbreviation, we define $\pi_{\text{ratio}}(y|x) = \frac{\pi_{\theta}(y|x)}{\pi_{\text{ref}}(y|x)}.$

Alternatives to DPO to align models to preference data in the offline, differentiable setting have been proposed. We discuss the most relevant to our work below and further in \cref{app:related_work}.

\paragraph{SLiC} \citet{zhao2022calibrating, zhao2023slichf} provides an alternative formulation for learning from pairwise preference data, either directly or on preference pairs sampled from the SFT policy. Their loss function is:

\begin{equation*}
\mathcal{L}_{\text{SLiC}}(\pi_{\theta}) = \mathbb{E}_{(x, y_w, y_l) \sim D, y_\text{ref}}
\left[\text{max}(0, \beta - \log \pi_{\theta}(y_w | x) + \log \pi_{\theta}(y_l | x)) 
- \lambda \log \pi_{\theta}(y_\text{ref} | x) \right]
\end{equation*}

where $y_\text{ref}$ is a reference target sequence and $\lambda$ is a scalar. The second term 
in the above is a cross-entropy loss and plays the role of regularisation towards the original model, but without needing an extra copy of the weights.

\paragraph{IPO} \citet{azar2023general} aims to understand the theoretical underpinnings of RLHF and DPO. They identify that DPO may be prone to overfitting in situations where the preference probability of the preferred over the dispreferred examples is close to 1. They propose an alternative form of pairwise preference loss---`Identity-PO (IPO)'. 
IPO tries to prevent overfitting to the preference dataset by penalising exceeding the preference margin beyond this regularised value. 
Conversely, we identify that DPO can lead to underfitting as well---even complete performance degradation.

\paragraph{Subsequent work}
Very recently, subsequent work has verified our main findings from \cref{sec:failure} and \cref{sec:dpop} on the existence of a failure mode of DPO and how to fix it \citep{pang2024iterative,feng2024towards,rafailov2024r}.
Pang et al.\ find that when fine-tuning on chain-of-thought reasoning tasks, adding a term similar to our \cref{eq:positive_dpo} is crucial to achieve strong performance \citep{pang2024iterative}.
Other work builds off of our theoretical insights by applying additional mathematical analysis, verifying our claims in \cref{sec:failure} that DPO can decrease the probability of preferred completions \citep{feng2024towards}.
Finally, Rafailov et al.\ analysed the same DPO failure mode phenomenon, showing that the likelihood of the preferred response should decrease during DPO training, when starting from a model that has undergone SFT \citep{rafailov2024r}.
We give additional discussion on related work in \cref{app:related_work}.

\section{Failure Mode of DPO} \label{sec:failure}

In this section, we take a step back and examine the DPO loss in \cref{eq:dpo}, specifically with an eye towards how it can reduce the probability of the preferred completion.
The loss is a function only of the difference in the log-ratios, which means that we can achieve a low loss value even if $\pi_{\text{ratio}}(y_w|x)$ is lowered below 1, as long as $ \pi_{\text{ratio}}(y_l|x)$ is also lowered sufficiently.
This implies that the log-likelihood of the preferred completions is reduced below the original log-likelihood from the reference model.

Why is this an issue? The original use-case of RLHF did not explicitly denote the preferred completions as being also \textit{satisfactory} completions (rather than just the preferred completion out of the two choices $y_w$ and $y_l$), and hence the DPO objective is a good modelling choice. 
However, since then, a large body of work has focused on distilling the knowledge of powerful models into smaller or weaker models, while also showing that doing so with RLHF/DPO outperforms SFT \citep{taori2023alpaca,tunstall2023zephyr,xu2023wizardlm,vicuna2023}. 
In this paradigm, it is often the case that in each pair of completions, the better of the two is indeed also a satisfactory completion. 
Furthermore, a new technique is to transform a standard labelled dataset into a pairwise preference dataset \citep{ivison2023camels,tunstall2023zephyr}, which also has the property that for each pair of completions, one is a satisfactory completion. 

\paragraph{Edit Distance 1}
While the above illustrates a hypothetical situation, now we provide a specific case in which DPO may cause a decrease in the probability of the better completion.
Consider the case of trying to improve a model's math or reasoning abilities by comparing a completion of ``2+2=4'' to ``2+2=5.''
This process creates a pair of preferred and dispreferred completions which have an edit (Hamming) distance of 1, i.e., all tokens in the completion are the same except for one.
In the following, we will explore how the location of the differing token impacts the computation of the DPO loss. For sake of argument, we will examine what happens when the differing token is the first token, though the argument also follows if it appears elsewhere. 

For preliminaries, consider two completions with an edit distance of 1 which differ at token $m$ with $1\leq m \leq K$, i.e., consider $y_w=(t_1, \dots, t_K)$ and $y_l=(t_1, \dots, t_{m-1}, t_m', t_{m+1}, \dots, t_K)$. Denote $y^{< r} = (t_1, \dots, t_{r-1})$ and $y^{\geq r} = (t_{r},\dots,t_K)$.
Assume that the vocabulary length of the LLM is $L$.
Let $s_i^{\{x\}}$ represent the probability of the $i$-th token in the model's vocabulary conditioned on the input  $x$.
While the LLM model parameters $\theta$ are numerous, we restrict our attention to the logits, $\theta_j$ with $j\in[L]$. 

The gradient of \cref{eq:dpo} with respect to $\theta$ is given by:
\begin{align*} 
&\nabla_{\theta} \mathcal{L}_{DPO} (\pi_{\theta}; \pi_{\text{ref}}) \propto -  \nabla_{\theta} \left[\log \pi_{\theta}(y_w | x) - \log \pi_{\theta}(y_l | x) \right]. \nonumber
\end{align*}

We note first that for any $m>1$, all tokens in positions from 1 to $m-1$ have no effect on the gradient, as for all $1\leq i\leq m-1$, $\pi_\theta(t_i|y_w^{<k},x)=\pi_\theta(t_i|y_l^{<k},x)$, causing these tokens' contribution to the gradient to cancel out.
Therefore, without loss of generality, assume $m=1$, i.e., $y_w$ and $y_l$ differ only at the first token.
Without loss of generality, we also assume that $t_k$ takes vocabulary position 1. Then we have the following  for each $k > 1$ (derivation in \cref{app:dpo_logits_gradients}):
\begin{align}\label{eq:dpo_gradient_by_logit}
    &\nabla_{\theta_j} \left[\log \pi_{\theta}(y_w | x) - \log \pi_{\theta}(y_l | x) \right]
    = s_j^{\{y_l^{<k},x\}} - s_j^{\{y_w^{<k},x\}}.
\end{align}

As we typically run DPO after SFT, the model is likely to be reasonably well optimised, so we should have $s_j^{\{y_w^{<k},x\}}\leq s_j^{\{y_l^{<k},x\}}$ for $j\neq 1$ and $s_1^{\{y_w^{<k},x\}}\geq s_1^{\{y_l^{<k},x\}}$. 
Therefore, while this analysis only extends to gradients with respect to the logits, we see that the gradient decreases logits corresponding to the correct token and increases logits corresponding to the incorrect tokens. Surprisingly, this suggests that under DPO, all locations in the sequence after the differing token will have reduced probability of emitting the correct token when compared to $\pi_{\text{ref}}$.
We give empirical evidence for this in \cref{sec:experiments} and \cref{fig:logprob_location}.

\section{DPOP} \label{sec:dpop}

Now, we introduce DPO-Positive (\methodShort{}), our proposed method to address the failure mode described in the previous section.
\methodShort{} adds the penalty term $\text{max}\left(0, \log \frac{ \pi_{\text{ref}}(y_w|x)}{ \pi_{\theta}(y_w|x)}\right)$ to the loss within the log-sigmoid to incentivise maintaining a high log-likelihood of the preferred completions. This penalty term is 0 when $\pi_{\text{ratio}}(y_w|x) \ge 1$ and increases as the ratio goes below 1.

The full DPOP loss function is thus:
\begin{align} \label{eq:positive_dpo}
\mathcal{L}_{\text{DPOP}}(\pi_{\theta}; \pi_{\text{ref}}) = -\mathbb{E}_{(x, y_w, y_l) \sim D} \bigg[ \log \sigma \bigg( \beta \bigg( \log \frac{\pi_{\theta}(y_w | x)}{\pi_{\text{ref}}(y_w | x)} &- \log \frac{\pi_{\theta}(y_l | x)}{\pi_{\text{ref}}(y_l | x)} \\ \nonumber
&-\lambda \cdot\text{max}\left(0, \log \frac{ \pi_{\text{ref}}(y_w|x)}{ \pi_{\theta}(y_w|x)}\right)\bigg) \bigg) \bigg]
\end{align}
where $\lambda > 0$ is a hyperparameter that can be tuned. This form of loss retains the property that we are fitting parameters on the preference data under the Bradley-Terry model. The implicit reward parameterisation is 

\begin{align*}
&\beta\cdot\log\frac{\pi_{\theta}(y|x)}{\pi_{\text{ref}}(y|x)} \text{ for } y = y_l,
&\beta\bigg[\log \frac{\pi_{\theta}(y_w|x)}{\pi_{\text{ref}}(y|x)} - \lambda\cdot \text{max}\left(0, \log \frac{ \pi_{\text{ref}}(y|x)}{ \pi_{\theta}(y|x)}\right) \bigg] \text{ for } y=y_w.
\end{align*}
By applying this optimisation pressure, the model can no longer minimise the loss by significantly reducing the log-likelihood of the dispreferred examples more than it reduces the log-likelihood of the preferred examples; it must also ensure that the log-likelihood of the preferred examples remains high relative to the log-likelihood under the reference model.

Now we show that \cref{eq:positive_dpo} mitigates the failure mode from the previous section.
Recall from \cref{sec:failure} that we focused on two completions, $y_w$ and $y_l$, which differ by one token at location $m=1$.
We showed in \cref{eq:dpo_gradient_by_logit} that for standard DPO, the gradient of the $k$-th token in the completions with respect to the $j$-th logit is $s_j^{\{y_l^{<k},x\}} - s_j^{\{y_w^{<k},x\}}$.
However, for \methodShort{}, if  $\pi_\text{ratio} < 1$, the gradients become (derivation in \cref{app:dpop_logits_gradients}):
\begin{align}
    \begin{split}
        \nabla_{\theta_j}{}&\bigg[\log \pi_{\theta}(t_k |y_w^{<k},x) -  \log \pi_{\theta}(t_k |y_l^{<k},x) +\lambda \cdot \log \pi_{\theta}(t_k |y_w^{<k},x)])\bigg]  \nonumber \\
    \end{split}\\
    &=\begin{cases}
        \lambda \left(1 - s_j^{\{y_w^{<k},x\}}\right) + s_j^{\{y_l^{<k},x\}} - s_j^{\{y_w^{<k},x\}} & \text{} i = j \nonumber \\
        -(\lambda + 1) s_j^{\{y_w^{<k},x\}} + s_j^{\{y_l^{<k},x\}} & \text{} i \neq j \nonumber \\
    \end{cases}
\end{align}
where $i$ is the vocabulary index of token $t_k$.
Since $s_j^{\{y_w^{<k},x\}} \le 1$, for the case $i=j$, the gradient is guaranteed to be positive for a large enough choice of $\lambda$. Similarly, for the case $i \neq j$, the gradient is guaranteed to be negative for a large enough $\lambda$ (as long as $s_j^{\{y_w^{<k},x\}} > 0$). This therefore fixes the issue we identified in DPO in \cref{sec:failure}.

\paragraph{Connection to Contrastive Loss}
While the main motivation for \methodShort{} is to avoid the failure mode described in \cref{sec:failure}, we also note its connection to \emph{contrastive loss}.
Contrastive learning  is a popular technique in areas such as computer vision for datasets of similar and dissimilar pairs \citep{oord2018representation,chen2020simple,he2020momentum}, and the loss function often uses a margin factor. 
\cref{eq:positive_dpo} can be viewed as similar to contrastive loss 
with margin
$m=\log\frac{1}{\pi_\text{ref}(y_w| x)}$.
We give further details in \cref{app:contrastive}.

\section{DPOP Datasets \& Experiments} \label{sec:experiments}

In this section, we empirically validate that the failure mode \textit{does} arise in practice and that \methodShort{} is able to mitigate the failure. We also show that even when the edit distance is large and DPO does not show degradation in performance, \methodShort{} can still outperform DPO on downstream task evaluation.

\subsection{Dataset Creation}

For our empirical analysis, we focus on the downstream tasks of \textbf{GSM8K}, \textbf{ARC}, and \textbf{HellaSwag}, and we introduce and release associated paired preference-ranked datasets.

\textbf{GSM8K} \citep{cobbe2021training}, a dataset of diverse grade school math word problems, has been adopted as a measure of the math and reasoning skills of LLMs \citep{chowdhery2023palm, touvron2023llama2, touvron2023llama, open-llm-leaderboard, eval-harness}.
We create a paired preference-ranked version of MetaMath \citep{yu2023metamath}, an extended version of the GSM8K training data \citep{an2023learning,yu2023metamath}. 
The correct completions in the MetaMath dataset consist of a series of steps which lead to the final answer. To create a dispreferred version, we randomly corrupt one of the results of an intermediate calculation. This dataset has a low (normalised) edit distance of 6.5\%.

\textbf{ARC} \citep{clark2018think} is a dataset that tests the level of understanding of science at  grade-school level. We focus specifically on ARC-Challenge, the more difficult of the two subsections of ARC, which has been widely adopted as a measure of LLM reasoning and world understanding \citep{chowdhery2023palm, touvron2023llama2, touvron2023llama, open-llm-leaderboard,cobbe2021training}.
The ARC-Challenge dataset consists of four choices of responses to each question, one of which is correct. To create a paired preference-ranked dataset, for each correct response in the training split, we create three pairs using each incorrect response. Due to the differences in the responses, this dataset has a high normalised edit distance of 90\%.

\textbf{HellaSwag} \citep{zellers2019hellaswag} is
a dataset containing commonsense inference questions known to be hard for LLMs.
Similar to ARC, each question has one correct completion and three incorrect completions, and so we create a paired preference-ranked dataset by creating three pairs for each correct response in the training split.
See \cref{app:datasets} for further details and documentation about our new datasets.

\begin{figure}[t]
    \centering
    \includegraphics[width=0.4\textwidth]{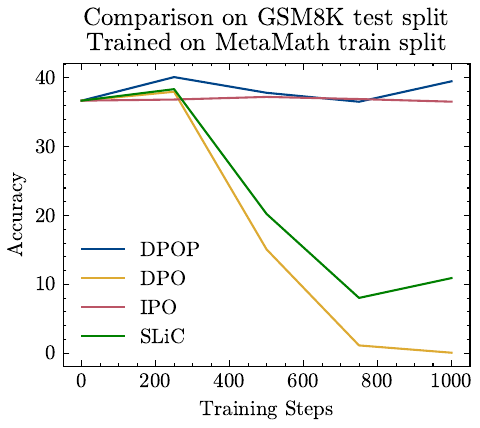}
    \includegraphics[width=0.4\textwidth]{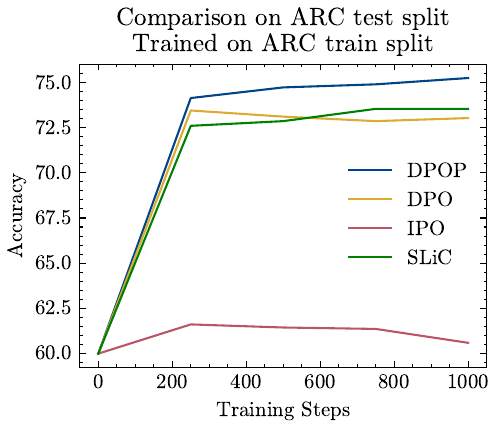}
    \caption{
    \textbf{Preference optimization comparisons on low (MetaMath) and high (ARC) edit distance datasets}.
    \methodShort{} outperforms DPO, IPO, and SLiC on both MetaMath (left) and ARC (right), whose normalized edit distances are 6.5\% and 90\%, respectively.
    Evaluation is performed on the test set of the datasets using the LLM Evaluation Harness.}
    \label{fig:dpo_vs_dpop}
\end{figure}

\subsection{Experiments} \label{subsec:results}

In this section, we compare training DPO, IPO, SLiC and \methodShort{} on the train splits of the datasets mentioned above and evaluate them on the corresponding tasks. We apply each preference-training method to the base model of Mistral7B \citep{jiang2023mistral}. We evaluate on the test sets of GSM8K and ARC with the LLM Evaluation Harness \citep{eval-harness}. Unless specified otherwise, we use values of $\beta = 0.3$ and $\lambda = 50$.

\paragraph{Loss function comparison} 
First, we compare DPO, IPO, SLiC and \methodShort{} when training on both MetaMath and ARC; see \cref{fig:dpo_vs_dpop}.
We find that when training on MetaMath, both DPO and SLiC catastrophically fail, while IPO does not improve performance. \methodShort{} is the only model to improve performance over the base model.
When training on ARC, which has a higher edit distance as described in the previous section, both DPO, SLiC and \methodShort{} are able to improve on the base model significantly; however, \methodShort{} performs best.

\begin{figure}[t]
    \centering
    \includegraphics[width=.32\linewidth]{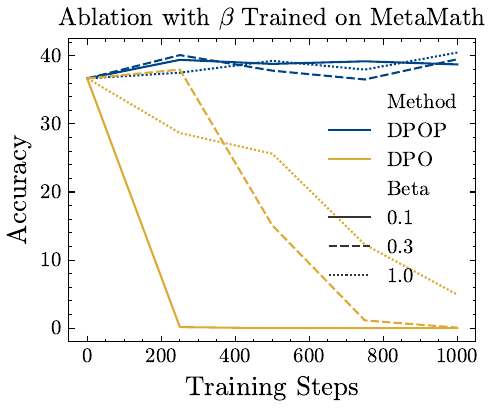}
    \includegraphics[width=.32\linewidth]{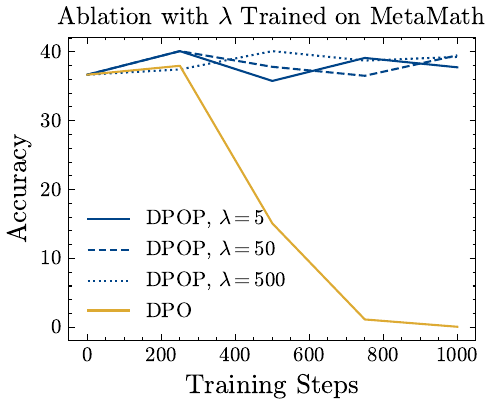}
    \includegraphics[width=.32\linewidth]{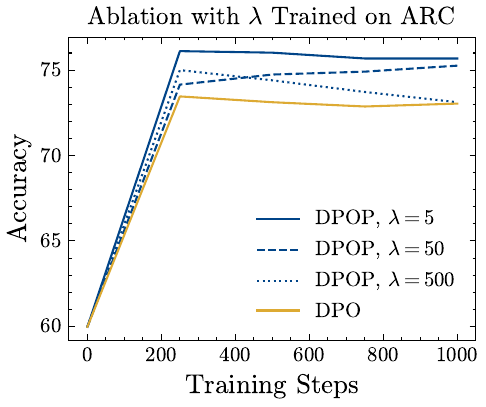}
    \caption{\textbf{Ablation studies:}    
    Evaluation of DPO vs.\ \methodShort{} for different values of $\beta$, on the MetaMath dataset (left), as well as for different values of $\lambda$, on the MetaMath (middle) and ARC (right) datasets.
    }
    \label{fig:lambda_ablation}
\end{figure}

\paragraph{Ablation studies for $\beta$ and $\lambda$} 
One potential hypothesis for how degradation of DPO on MetaMath could be prevented is by modifying the strength of the regularisation parameter, $\beta$. We test $\beta\in\{0.1,0.3,1.0\}$, and although a larger $\beta$ does induce a slower decrease, the performance with DPO still plummets, while  \methodShort{} shows strong and consistent performance with different values of $\beta$ (see \cref{fig:lambda_ablation}).
Furthermore, we conduct an ablation study over the value of $\lambda$ (a hyperparameter unique to \methodShort) to determine the sensitivity of performance to tuning this value precisely. We test $\lambda\in\{5,50,500\}$ and find that performance on MetaMath and ARC is relatively unaffected by the choice of $\lambda$. See \cref{fig:lambda_ablation}.

\paragraph{Token-level analysis} 
Recall that in \cref{sec:failure}, we gave theoretical motivations for why DPO is likely to perform poorly on low-edit distance datasets. We now analyse the log-probabilities of the trained models at the token level on the MetaMath dataset over 900 samples to empirically support our arguments. Let us denote the index of the first token that is different between the preferred and dispreferred completion by $m$. 

\begin{figure}[t]
    \centering
    \includegraphics[width=0.49\textwidth]{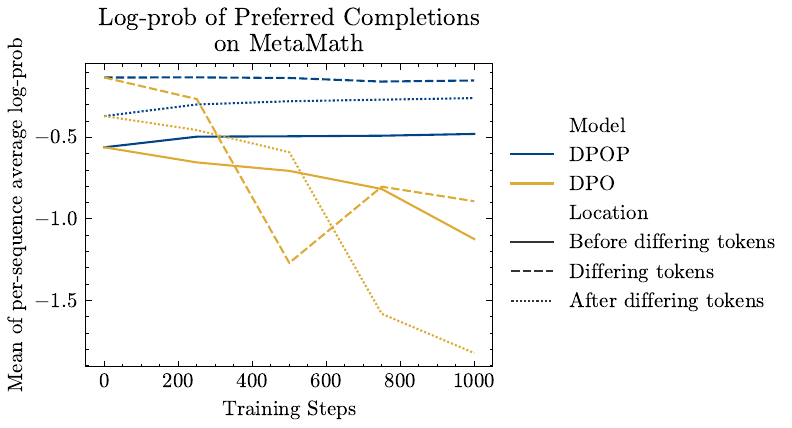}
    \includegraphics[width=0.49\textwidth]{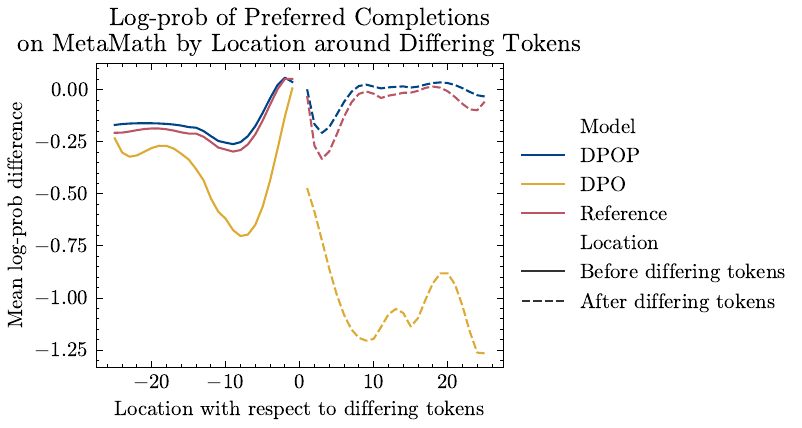}
    \caption{
    \textbf{DPO fails to train on low edit-distance pairs, yet \methodShort{} performs well.}
    Left: average log-probs for 900 randomly-sampled preferred train set completions on MetaMath across training steps. For DPO, the log-probs decrease throughout training.
    Right: average log-prob difference for preferred completions on MetaMath by location around differing tokens, after 1000 training steps. Log-prob `difference' signifies that each model's plot has been adjusted to have 0 log-prob at location -1; all other log-probs are shown relative to this value.
    For DPO, there is a significant decrease after the differing tokens, while \methodShort{} avoids this issue. The log-probs of the LLM prior to application of \methodShort{} or DPO is also included for reference.
        }
    \label{fig:logprob_location}
\end{figure}

We suggested that  $\pi_\theta(y^{\geq r}\mid x,y^{< r})$ for $r > m$ will have `wrong-way' gradient updates and therefore decrease. We find this is indeed the case---the average log-prob after training of tokens after $m$ is $-0.37$ for the reference model and $-0.26$ for \methodShort{}, but $-1.82$ for DPO on the preferred completions (see (\cref{fig:logprob_location}) (left)). Perhaps most instructively, for both the reference model and \methodShort{}, in \cref{fig:logprob_location} (right), we see that tokens after the edit indices show higher log-likelihood than those before the edit indices---this is indicative of well-behaved language modelling, with lower perplexity as more tokens are added to the context. By contrast, DPO shows the opposite pattern---with log-likelihood actually reducing after the edit indices. This is indicative of a deeper breakdown in language modelling, which we believe is facilitated by the wrong-way gradient we outlined in \cref{sec:failure}. Finally, we are also able to substantiate our assumption from \cref{sec:failure} that $s_1^{\{y_w^{<k},x\}}\geq s_1^{\{y_l^{<k},x\}}$; we find from our analysis that for the baseline model, the tokens after the edit have an average log-likelihood of $-0.37$ on the preferred completion, but this drops to $-0.86$ on the dispreferred completion. 

In \cref{app:experiments}, we present additional results on ARC comparing the averaged log-probs of DPO and \methodShort{} on the preferred completion during training; see Appendix \cref{fig:dpo_failures_arc}. \methodShort{} once again demonstrates higher log-probs than DPO.

\section{\model} \label{sec:smaug}

In this section, we introduce the \model{} series of models.
We train models for 7B, 34B and 72B parameter sizes using DPOP.
We use the 7B class for a direct comparison of \methodShort{} vs.\ DPO, including testing the generalisability of the improvement in model performance on benchmarks very different from the training data, such as MT-Bench \citep{zheng2023judging}.
Due to the computational resource requirements involved in training the larger model sizes, we only perform \methodShort{} on 34B and 72B and compare to other models on the HuggingFace Open LLM Leaderboard.
Also due to computational expense, we do not perform any further hyperparameter tuning.

\paragraph{Datasets} In this section, unless otherwise noted, we train on a mix of six pair preference datasets: our MetaMath DPO, ARC DPO, and HellaSwag DPO datasets described in \cref{sec:experiments}, the ORCA DPO dataset \citep{orcaDpoPairs}, the Truthy DPO dataset \citep{truthyDpo}, and the UltraFeedback Binarized dataset \citep{ultrafeedbackBinarizedCleaned}.
We run these experiments with 8 H100 GPUs (each with 80GB), using the \texttt{transformers} repository  \citep{wolf-etal-2020-transformers} for general model loading and infrastructure, and the TRL repository \citep{vonwerra2022trl} for running DPOP. 
We set $\beta=0.3$, $\lambda=50$, a learning rate of $5\times10^{-5}$, and the AdamW optimizer \citep{loshchilov2017decoupled}, and we run 1000 steps for all DPO and \methodShort{} routines.

\subsection{DPOP vs.\ DPO Ablation on 7B}

First we apply the above recipe on Llama-2-Chat \citep{touvron2023llama2}.
Since this model has already undergone instruction fine-tuning, we perform DPO and \methodShort{} directly using the datasets described in the previous section. We evaluate the Llama-2 finetunes on MT-Bench \citep{zheng2023judging}
This benchmark tests across multiple categories of LLM performance (for example: writing, roleplay, coding and math).

\paragraph{MT-Bench} We evaluate MT-Bench on Llama-2-7B-Chat finetuned with DPO and \methodShort{}. We run MT-Bench with the Llama-2 conversation template. We perform 10 trials due to the inherent stochasticity of the benchmark, computing mean and standard deviation. Across the 10 trials, DPO achieves a first turn score of $7.032\pm 0.043$ whilst \methodShort{} scores $7.292\pm 0.037$, a significant improvement.

MT-Bench tests across eight diverse categories and is significantly different from our fine-tuning data. Furthermore, MT-Bench has been shown to have performance correlated to human preferences of LLM rankings \citep{zheng2023judging, rafailov2023direct, arenahard2024}.  The outperformance of \methodShort{} vs.\ DPO in this controlled like-for-like setting is therefore an indication of the improvement gained by utilising \methodShort{} over DPO on new tasks not in the training set, and also an indication of general model quality.

\subsection{\model-34B and \model-72B}

\model-34B is a modified version of the base model 
Bagel-34B-v0.2 \citep{bagel34B}, which itself is a SFT version of Yi-34B-200k \citep{Yi34B200K}. We first take Bagel-34B-v0.2 and perform a SFT fine-tune using a combination of three datasets: MetaMath \citep{yu2023metamath}, ORCA-Chat \citep{orcaChat}, and the ShareGPT dataset \citep{ShareGPTVicunaunfiltered}. We then apply \methodShort{} using the methodology and datasets described above.

For 72B, we start from MoMo-72b-lora-1.8.7-DPO \citep{MoMo72B}, which itself is a fine-tune of Qwen-72B \citep{bai2023qwen}.
MoMo-72b-lora-1.8.7-DPO has already undergone SFT, so we simply run the \methodShort{} routines as in \model-34B.
The total training time is 144 hours.

\paragraph{HuggingFace Open LLM Leaderboard}
We evaluate using the HuggingFace Open LLM Leaderboard \citep{open-llm-leaderboard,eval-harness}, a widely-used benchmark suite that aggregates six popular benchmarks: 
ARC \citep{clark2018think},
GSM8K \citep{cobbe2021training}, 
HellaSwag \citep{zellers2019hellaswag}, 
MMLU \citep{hendrycks2021measuring}, 
TruthfulQA \citep{lin2022truthfulqa}, and
WinoGrande \citep{sakaguchi2020winogrande}.
We evaluate directly in HuggingFace, which uses the LLM Evaluation Harness \citep{eval-harness}. It performs few-shot prompting on the test sets of these tasks. 
We compare \model-72B to the evaluation scores of the top five open-weight LLMs according to the HuggingFace Open LLM Leaderboard \citep{open-llm-leaderboard,eval-harness} as of March 2024; see \cref{tab:leaderboard}.
\model-72B achieves an average accuracy of 80.48\%, becoming the first open-source LLM to surpass an average accuracy of 80\% and improving by nearly 2\% over the second-best open-source model.
\model-34B also achieves best-in-its-class performance compared to other models of similar or smaller size (see \cref{app:experiments}).

\paragraph{MT-Bench}
Next, we evaluate again using MT-Bench \citep{zheng2023judging}.
We run MT-Bench with the Llama-2 conversation template \citep{touvron2023llama2}. 
See Appendix \cref{tab:mtbench} for a comparison with state-of-the-art LLMs according to Arena Elo as of March 2024. 
\model-72B achieves the top MMLU score and third-best MT-bench score out of the open-source models. 
In \cref{app:experiments}, we give examples of \model-72B completions to MT-Bench questions.

\subsection{Contamination check}
Since LLMs train on wide swaths of the internet, data contamination, i.e., evaluating on examples that are included in the training data, is a growing concern.
Data contamination remains notoriously challenging to measure and mitigate, even with partial attempts like controlled experimentation of training data, canary strings, or embedding similarities \citep{roberts2024cutoff,jain2024livecodebench,srivastava2023imitation}.

While there is no perfect tool to check for data contamination, we use an open-source contamination checker \citep{shi2023detect}
to compare the contamination of our model to other open-source models on training datasets of the Huggingface Open LLM Leaderboard. On ARC, TruthfulQA, and GSM8K, we find that \model-72B achieves scores that are similar to MoMo-72B-lora-1.8.7-DPO (the starting point of \model-72B), as well as Llama-2-70B.
See the full details in \cref{app:contamination}.

\begin{table*}[t]
\caption{Evaluation of the top open-weight models on the HuggingFace Open LLM Leaderboard as of March 2024.
See \cref{tab:leaderboard_extension} for an extended comparison. 
}
\label{tab:leaderboard}
\resizebox{\textwidth}{!}{%
\centering
\begin{tabular}{lrrrrrrrr}
\toprule
Model & Size &  Avg.\ & ARC & HellaSwag & MMLU & TruthfulQA & Winogrande & GSM8K \\
\midrule
\textbf{\model-72B (Ours)} & 72B+ & \textbf{80.48} & \textbf{76.02} & 89.27 & \textbf{77.15} &  76.67 & 85.08 & \textbf{78.70} \\
MoMo-72B-lora-1.8.7-DPO & 72B+ &  78.55 & 70.82 & 85.96 & 77.13 & 74.71 & 84.06 & 78.62 \\
TomGrc\_FusionNet\_34Bx2\_MoE\_v0.1\_DPO\_f16 & 72B+ & 77.91 & 74.06 & 86.74 & 76.65 & 72.24 & 83.35 & 74.45 \\
TomGrc\_FusionNet\_34Bx2\_MoE\_v0.1\_full\_linear\_DPO & 72B+ &  77.52 & 74.06 & 86.67 & 76.69 & 71.32 & 83.43 & 72.93 \\
Truthful\_DPO\_TomGrc\_FusionNet\_7Bx2\_MoE\_13B & 72B+ &  77.44 & 74.91 & \textbf{89.30} & 64.67 & \textbf{78.02} & \textbf{88.24} & 69.52 \\
\bottomrule
\end{tabular}
}
\end{table*}

\section{Conclusions and Limitations} \label{sec:conclusion}

In this work, we presented new theoretical findings on a severe failure mode of DPO, in which fine-tuning causes the probability of the preferred examples to be reduced. 
We then presented an empirical token-level analysis that matches our theoretical findings on popular datasets. 
In order to mitigate this issue, we devised a new technique, \methodShort{}, which we show overcomes the failure mode of DPO --- and can outperform DPO even outside this failure mode.
By fine tuning with \methodShort{} on our new pairwise preference versions of ARC, HellaSwag, and MetaMath, we create a new LLM that achieves state-of-the-art performance. In particular, we present the first open-weights model to surpass an average accuracy of 80\% on the HuggingFace Open LLM Leaderboard, and we also show that DPOP significantly outperforms DPO in an apples-to-apples comparison on MT-Bench, which is independent of the fine-tuning data.

In the future, creating pairwise preference-based versions of other datasets, and running \methodShort{} with these datasets, could push the abilities of open-source LLMs closer to the performance of proprietary models such as GPT-4 \citep{gpt4}, especially when tuned for specific downstream tasks.
Furthermore, using \methodShort{} on additional mathematical datasets is an exciting area for future work, as it has the potential to further advance LLMs' abilites in mathematical reasoning.

\paragraph{Limitations}
While our work gives theoretical and empirical evidence on a failure mode of DPO and a proposed solution, it still has limitations.
First, we were unable to run a full ablation study on our 72B model. Running multiple fine-tuning experiments on a 72B model is infeasible, as each one can take over five days to complete. Therefore, we assume that our ablations on smaller models still hold up at scale.
Furthermore, while we expect \methodShort{} to achieve strong performance on any preference dataset, especially those with small edit distance, we have only demonstrated its performance on six English-language datasets. We hope that future work can verify its effectiveness on more datasets, in particular on non-English datasets.

\bibliography{main}

\clearpage
\appendix

\section{Related Work Continued} \label{app:related_work}

In this section, we continue our discussion of related work from \cref{sec:background}.

\paragraph{AFT} \citet{wang2023making} seek to align LLMs to correctly `score' (in terms of perplexity) their own generations. They do so by generating multiple chain-of-thought \citep{wei2023chainofthought} responses to each prompt, which they categorise as preferred or dispreferred according to whether they answer the question correctly. Their proposed `Alignment Fine-Tuning (AFT)' paradigm adds an alignment objective $\mathcal{L}^{*}_{A}$ to the standard fine tuning loss, defined as
\begin{align*} 
&\mathcal{L}^{*}_A (\pi_\theta) = \log \bigg[ 1 +
\sum_{y_w \in \mathcal{G}_W} \sum_{y_l \in \mathcal{G}_L} e^{(\log \pi_\theta(y_l | x) - \log \pi_\theta(y_w | x))} \bigg]
\end{align*}
where $\mathcal{G}_p$ is the set of preferred examples and $\mathcal{G}_n$ is the set of dispreferred examples. By minimising $\mathcal{L}^{*}_A$, the log-likelihoods of preferred examples are encouraged to be larger than the log-likelihoods of dispreferred examples, akin to DPO. However, \citet{wang2023making} takes an opposing motivation to us: they are particularly concerned with the issue of the log-likelihoods of dispreferred examples being pushed down too significantly 

Our work differs from AFT in three key points. First, although \citet{wang2023making} discusses DPO in the appendix, they do not show how their approach would extend to a reformulation of its objective; they also focus their experiments solely on supervised fine-tuning.
Next, we use a different constraint mechanism---theirs is a soft margin constraint on the log-probability distance of the dispreferred example from the preferred example, while ours is a soft penalty for deviating from a reference model.  Finally, they are focused specifically on the case of self-generated LLM CoT responses and calibrating the LLM's perplexity of its own responses.

\paragraph{HALO} \citet{ethayarajh2023halos} seek to understand alignment methods, including DPO, in the context of `Human-Centred Loss Functions (HALOs)'. By drawing an equivalence between the alignment methods and the work of \citet{TverskyKahneman1992} in prospect theory, they adapt the `human value function' in that paper to the LLM setting:
\begin{equation*}
L_{KTO}(\pi_{\theta}, \pi_{\text{ref}}) = \mathbb{E}_{x, y \sim D} \left[ w(y) (1 - \hat{h}(x, y; \beta)) \right]
\end{equation*}
where they define $g(x, y; \beta)$ as
\begin{equation*}
 \beta \log \frac{\pi_{\theta}(y|x)}{\pi_{\text{ref}}(y|x)} - \mathbb{E}_{x' \sim D} \left[ \beta \text{KL} (\pi_{\theta} || \pi_{\text{ref}}) \right] 
 \end{equation*}
 and
 \begin{align*}
 h(x, y; \beta) &= 
\begin{cases}
\sigma(g(x, y; \beta)) & \text{if } y \sim y_{\text{w}}|x \\
\sigma(-g(x, y; \beta)) & \text{if } y \sim y_{\text{l}}|x
\end{cases} \\
w(y) &= 
\begin{cases}
\lambda_{D} & \text{if } y \sim y_{\text{w}}|x \\
\lambda_{U} & \text{if } y \sim y_{\text{l}}|x
\end{cases}
\end{align*}

The major difference of this approach with DPO is that it does not require paired preference data. The above loss function can be used for any dataset as long as the labels are individually marked as positive or negative.

\paragraph{CPO} Very recently, concurrent work \citep{xu2024contrastive} proposes adding a new term to the DPO loss function in order to allow DPO to become better at rejecting `worse' completions that are good quality but not perfect.
Specifically, they include the term

\begin{equation*}
\E_{(x, y_w, y_l) \sim D} \left[\log\pi_\theta(y_w\mid x)\right].
\end{equation*}

While similar, their work uses a different loss function with different motivation, and furthermore they only considered machine translation models up to 13B parameters.

\paragraph{Subsequent work, extended discussion}
As described in \cref{sec:background}, very recently, subsequent work has verified our main findings from \cref{sec:failure} and \cref{sec:dpop} on the existence of a failure mode of DPO and how to fix it \citep{pang2024iterative,feng2024towards,rafailov2024r}.
We give additional discussion here.

Pang et al.\ consider fine-tuning on chain-of-thought reasoning tasks using pairwise preference data. They find that including an additional positive log-likelihood term is crucial for their reasoning tasks. Their final loss is similar to our \cref{eq:positive_dpo}, but the additional term is outside the log sigmoid \citep{pang2024iterative}.

Feng et al.\ give additional mathematical analyses of the gradient vector fields of the DPO loss. They come to the conclusion that the DPO loss function decreases the probability of producing dispreferred completions at a faster rate than it increases the probability of producing preferred completions \citep{feng2024towards}.

Finally, Rafailov et al.\ also analysed the ``phenomena in which the likelihood of the chosen responses actually decrease over time [in DPO]'' by examining the expected log ratio of a completion under a model being optimised to a reference model. By showing that this is equivalent to the KL divergence, which is necessarily non-negative, they find that ``the likelihood of the chosen response should decrease in the process of DPO training'' when starting from a model that has undergone SFT on the preferred completions \citep{rafailov2024r}.

\section{Derivation of logit gradients} \label{app:logits_gradients}

\subsection{Derivation for DPO}\label{app:dpo_logits_gradients}

Consider two completions of length $K$ with edit (Hamming) distance of 1 which differ at token $m$ with $1\leq m \leq K$.
Put $y_w=(t_1, \dots, t_K)$ and $y_l=(t_1, \dots, t_{m-1}, t_m', t_{m+1}, \dots, t_K)$. Put $y^{< r} = (t_1, \dots, t_{r-1})$ and $y^{\geq r} = (t_{r},\dots,t_K)$.
Note that the derivative of \cref{eq:dpo} with respect to $\theta$ is proportional to:
\begin{align*}
&\nabla_{\theta} \mathcal{L}_{DPO} (\pi_{\theta}; \pi_{\text{ref}}) \propto - \left[ \nabla_{\theta} \log \pi_{\theta}(y_w | x) - \nabla_{\theta} \log \pi_{\theta}(y_l | x) \right]
\end{align*}
By the chain rule, we have
\[
    \pi_\theta(y|x) = \prod_{k=1}^K \pi_\theta(t_k | y^{<k},x)
\]
and by using logarithm identities
\[
    \log\pi_\theta(y|x) = \sum_{k=1}^K \log\pi_\theta(t_k | y^{<k},x).
\]
When we substitute this into $\nabla_{\theta} \mathcal{L}_{DPO}$, we have:

\begin{align}\label{eq:sumoflogs}
    \nabla_{\theta} \mathcal{L}_{DPO} (\pi_{\theta}; \pi_{\text{ref}}) \propto
&- \sum_{k=1}^K \nabla_{\theta} \left[ \log \pi_{\theta}(t_k | y_w^{<k}, x) - \log \pi_{\theta}(t_k | y_l^{<k}, x) \right]
\end{align}

Assume that the vocabulary length of the LLM is $L$. While the LLM model parameters $\theta$ are numerous, let us restrict our attention to just the logits, which we denote as $\theta_j$ with $j\in[L]$ and are the input to the softmax. We further denote the softmax probabilities by ${s^{\{x\}}}$, so that $s_i^{\{x\}}$ represents the probability of the $i$-th token in the model's vocabulary given the input context $x$.

It is then a standard result of the softmax function that: 
\begin{align}\label{eq:partiallogs}
    \frac{\partial}{\partial\theta_j} \log s_i^{\{x\}} = 1\{i=j\} - s_j^{\{x\}}.
\end{align}

Consider the case where $y_w$ and $y_l$ differ only at the first token, i.e., $m=1$, and without loss of generality, assume that $t_k$ takes vocabulary position 1.  In this instance, for each term of \cref{eq:sumoflogs} where $k > 1$ we have\footnote{Note that the preceding negative sign is dropped as it is the gradient of a loss function and so during optimisation we take steps in the opposite direction, so as to minimise the loss.}:
\begin{align}
    \nabla_{\theta_j} \log \pi_{\theta}(t_k |y_w^{<k},x) - \nabla_{\theta_j} \log \pi_{\theta}(t_k |y_l^{<k},x) &= 1\{1=j\} - s_j^{\{y_w^{<k},x\}} - (1\{1=j\} - s_j^{\{y_l^{<k},x\}}) \nonumber \\
    &= s_j^{\{y_l^{<k},x\}} - s_j^{\{y_w^{<k},x\}}.
\end{align}
As we typically run DPO after SFT the model is likely to be reasonably well optimised, so we should have $s_j^{\{y_w^{<k},x\}}\leq s_j^{\{y_l^{<k},x\}}$ for $j\neq 1$ and $s_1^{\{y_w^{<k},x\}}\geq s_1^{\{y_l^{<k},x\}}$. 
Therefore, while this analysis only extends to gradients with respect to the logits, we see that the gradient vector is decreasing in the correct logit dimension and increasing in the wrong logit dimensions. In particular, this derivation suggests that under DPO, all tokens that follow a difference at $m$ at any point should have reduced probability of emitting the correct token when compared to $\pi_{\text{ref}}$. 

\subsection{Derivation for \methodShort{}}
\label{app:dpop_logits_gradients}

We can follow a similar line of reasoning for calculating $\nabla_\theta \mathcal{L}_{DPOP}$ with respect to its logits which we denote again by $\theta_j$.

Again taking token position $k$ for illustrative purposes and assuming that token $t_k$ has vocabulary position $i$, in the case when $\pi_{\text{ratio}}(y|x) < 1$,
\begin{align}
    \begin{split}
        \nabla_{\theta_j}{}&[\log \pi_{\theta}(t_k |y_w^{<k},x) - \log \pi_{\theta}(t_k |y_l^{<k},x) \nonumber -\lambda \text{max}(0, \log \frac{ \pi_{\text{ref}}(t_k |y_w^{<k},x)}{ \pi_{\theta}(t_k |y_w^{<k},x)})] \nonumber \\
        &=\nabla_{\theta_j} [\log \pi_{\theta}(t_k |y_w^{<k},x) - \log \pi_{\theta}(t_k |y_l^{<k},x)
        -\lambda \log \frac{ \pi_{\text{ref}}(t_k |y_w^{<k},x)}{ \pi_{\theta}(t_k |y_w^{<k},x)}]  \nonumber \\        
        &= \nabla_{\theta_j} [\log \pi_{\theta}(t_k |y_w^{<k},x) - \log \pi_{\theta}(t_k |y_l^{<k},x) 
        +\lambda \log \pi_{\theta}(t_k |y_w^{<k},x)]  \nonumber \\
        &=(1+\lambda)\nabla_{\theta_j} \log \pi_{\theta}(t_k |y_w^{<k},x)
        - \nabla_{\theta_j} \log \pi_{\theta}(t_k |y_l^{<k},x) \nonumber \\        
        &= (1+\lambda)(1\{i=j\} - s_j^{\{y_w^{<k},x\}}) 
        - (1\{i=j\} - s_j^{\{y_l^{<k},x\}}) \nonumber \\      
        &=\begin{cases}
            \lambda (1 - s_j^{\{y_w^{<k},x\}}) + s_j^{\{y_l^{<k},x\}} - s_j^{\{y_w^{<k},x\}} & \text{} i = j \nonumber \\
            -(\lambda + 1) s_j^{\{y_w^{<k},x\}} + s_j^{\{y_l^{<k},x\}} & \text{} i \neq j \nonumber \\
        \end{cases}
    \end{split}
\end{align}

Since $s_j^{\{y_w^{<k},x\}} \le 1$, for the case $i=j$, the gradient is guaranteed to be positive for a large enough choice of $\lambda$. Similarly, for the case $i \neq j$, the gradient is guaranteed to be negative for a large enough $\lambda$ (as long as $s_j^{\{y_w^{<k},x\}} > 0$). This therefore fixes the issue in DPO from \cref{sec:failure}.

In the case when $\pi_{\text{ratio}}(y|x) \geq 1$, then we have the standard gradient from $\mathcal{L}_{\text{DPO}}$.

\section{Motivation: Contrastive Loss} \label{app:contrastive}

While the main motivation for \methodShort{} is to avoid the failure mode described in \cref{sec:failure}, we also note its connection to \emph{contrastive loss}.
Contrastive learning is widely used \citep{wang2021understanding,wang2020understanding,saunshi2019theoretical,oord2018representation,chen2020simple,he2020momentum}, often for embedding learning applications. 
The contrastive loss formulation typically includes two main terms: one encouraging the proximity of analogous inputs, the other encouraging the divergence of distinct classifiable data.

Moreover, the introduction of a margin appended to one of these terms often ensures a more stable training process. This margin serves as an indicator of indifference to point displacement once a specific value threshold is exceeded. The margin, when attached to the similar points term, establishes a minimum threshold beyond which we do not care about pulling the points closer. Alternatively, if added on the dissimilar points term, the margin sets a maximum threshold. 

We show that the DPO loss is structured such that learning the probabilities during DPO training are equivalent to learning the embeddings in a contrastive loss formulation. However, the standard DPO only uses the term computing distance between dissimilar points, and does not include the similar points term or the margin. Consequently, it is predictable that traditional DPO's inefficiencies mirror the known shortcomings of contrastive training when one constituent term is absent. \methodShort{}, our refined DPO formulation, fixes this by adding the absent term and the margin.

Contrastive loss is defined in \cite{contrastiveLoss}. If we keep the margin in the similar points terms, it can be written as follows:
\begin{align*} \mathcal{L}_{\text{Cont}} = & - \sum_{\forall (i,j) \epsilon \mathcal{P}_d}\mathcal{D}(y_i,y_j) + \lambda\sum_{\forall (i,j) \epsilon \mathcal{P}_s}\min(\mathcal{D}(y_i,y_j)-m,0)
\end{align*}

Recall that the standard DPO loss (\cref{eq:dpo}) is as follows:
\begin{align*}
&\mathcal{L}_{\text{DPO}}(\pi_{\theta}; \pi_{\text{ref}}) = -\mathbb{E}_{(x, y_w, y_l) \sim D} 
\left[ \log \sigma \left( \beta \log \frac{\pi_{\theta}(y_w | x)}{\pi_{\text{ref}}(y_w | x)} - \beta \log \frac{\pi_{\theta}(y_l | x)}{\pi_{\text{ref}}(y_l | x)} \right) \right] \nonumber
\end{align*}

Say we designate an embedding function $\mathcal{H}$:
\begin{align*}
\mathcal{H}(y|x) = \frac{\pi_{\theta}(y | x)}{\pi_{\text{ref}}(y | x)}.
\end{align*}

And we define a distance function $\mathcal{D}$ as follows:
\begin{align*}
\mathcal{D}(p_i,p_j) = \log[\mathcal{H}(p_i)] - \log[\mathcal{H}(p_j)].
\end{align*}

The standard DPO only has the dissimilar points term under the analogy of the contrastive loss formulation. For more robust training we accommodate for the similar embeddings term. 
We use the concept of anchor points or embeddings for both positive and negative points as in triplet loss \cite{triplet_loss}. These points are known ideal embeddings we want our points to achieve. They carry probabilities of 1 and 0 respectively in our equivalence depending on whether they are preferred or dispreferred samples.

\begin{align*}
\mathcal{H}_{p}^*(y|x) &= \frac{1}{\pi_{\text{ref}}(y | x)}\\ \mathcal{H}_{n}^*(y|x) &= \left.\frac{\epsilon }{\pi_{\text{ref}}(y | x)}\right\vert_{\epsilon \to 0}
\end{align*}

The \methodShort{} loss can thus be formulated as:
\begin{align*}
   \mathcal{L}_{\text{DPOP}} &= \left[ \log(\mathcal{H}(y_w|x)) - \log(\mathcal{H}(y_l|x))\right ] + \lambda\bigg[\min\bigg(\log(\mathcal{H}^*_p(y_w|x))- \log(\mathcal{H}(y_w|x)) - m,0 \bigg ) \\ 
   &\hspace{5.425cm}+\min\left(\log(\mathcal{H}^*_n(y_l|x)) - \log(\mathcal{H}(y_l|x)) - m,0 \right ) \bigg]\\
 &= \bigg[ \log \frac{\pi_{\theta}(y_w | x)}{\pi_{\text{ref}}(y_w | x)} 
- \log \frac{\pi_{\theta}(y_l | x)}{\pi_{\text{ref}}(y_l| x)} \bigg]+ \lambda\bigg[\min\bigg(\log \frac{1}{\pi_{\text{ref}}(y_w | x)} - \log \frac{\pi_{\theta}(y_w | x)}{\pi_{\text{ref}}(y_w | x)} - m,0\bigg)\bigg] \\ 
 &\hspace{5.825cm}+ \lambda\bigg[\min\bigg(\log \frac{\epsilon }{\pi_{\text{ref}}(y_l | x)} - \log \frac{\pi_{\theta}(y_l | x)}{\pi_{\text{ref}}(y_l | x)} - m,0\bigg)\bigg]
\end{align*}

If we set the margin $m = \log \frac{1}{\pi_{\text{ref}}(y_w | x)} $, the second term is:
\begin{align*}
     = - \lambda\left[\max\left(\log \frac{\pi_{\theta}(y_w | x)}{\pi_{\text{ref}}(y_w | x)},0\right)\right]
\end{align*}
This choice of margin is mathematically equivalent to choosing a threshold which ensures the similarity term only contributes to the loss when the learned model performs worse on the preferred response than the base model.

We can ignore the third term during training for two primary reasons. First, it is trying to push the log probability of negative samples to negative infinity which may be unstable during training. Second, in essence, it negatively impacts the likelihood of the negative samples. However, given our objective of extending the distance between positive and negative samples without diminishing the likelihood of positives, sacrificing this signal is acceptable. In the worst-case scenario, while the probability of negatives may be increased, the consequent increase in the probability of positives is ensured by the first two terms, thus rendering the sacrifice acceptable.

\section{Details of Datasets} \label{app:datasets}

In this section, we give more details and examples of the three paired preference versions of datasets we release.
We release all datasets under the Apache 2.0 License.

\paragraph{MetaMath} GSM8K \citep{cobbe2021training} is a dataset of diverse grade school maths word problems, which has been commonly adopted as a measure of the math and reasoning skills of LLMs \citep{chowdhery2023palm, touvron2023llama2, touvron2023llama, open-llm-leaderboard, eval-harness}.
The MetaMath dataset is partitioned into queries and responses, where the query is a question involving mathematical calculation or reasoning, and the response is a logical series of steps and calculations that culminate in a final answer. To construct our paired-preference version of MetaMath, we take the queries as prompts $x$ and the responses as the preferred completions $y_w$. We create $y_l$ by modifying the response by randomly corrupting one of the results of an intermediate calculation so that it is incorrect; however, we leave the answer untouched. 

An example is as follows:
$x$: ``What is the total cost of purchasing equipment for all sixteen players on the football team, considering that each player requires a \$25 jersey, a \$15.20 pair of shorts, and a pair of socks priced at \$6.80?"
$y_w$: ``Each player requires a \$25 jersey, a \$15.20 pair of shorts, and a pair of socks priced at \$6.80. So the total cost for each player is \$25 + \$15.20 + \$6.80 = \$\textbf{47}. Since there are sixteen players on the football team, the total cost for all of them is 16 * \$47 = \$752. \#\#\#\# 752 The answer is: 752"
$y_l$: ``Each player requires a \$25 jersey, a \$15.20 pair of shorts, and a pair of socks priced at \$6.80. So the total cost for each player is \$25 + \$15.20 + \$6.80 = \$\textbf{52}. Since there are sixteen players on the football team, the total cost for all of them is 16 * \$47 = \$752. \#\#\#\# 752 The answer is: 752"

The dataset contains 393\,999 training examples and 1\,000 evaluation examples.
Our motivation in building this dataset is to align models towards being precise in intermediate calculations. This dataset has low edit distance -- the normalised edit distance is approximately 6.5\%.

\paragraph{ARC}
ARC \citep{clark2018think} is a dataset that tests the level of understanding of science at approximately grade-school level. We focus specifically on the `Challenge' subsection of ARC, the more difficult of the two subsections, which has been widely adopted as a measure of LLM reasoning and world understanding \citep{chowdhery2023palm, touvron2023llama2, touvron2023llama, open-llm-leaderboard, eval-harness,cobbe2021training}.
We create a paired preference-ranked dataset from the train split of ARC-Challenge. The dataset is partitioned into questions which we take as our prompts $x$, and four choices of responses to each question of which only one is the correct answer. The correct response is taken as $y_w$ and the incorrect responses are taken to be $y_l$; as there are three incorrect responses for every prompt, we repeat $y_w$ multiple times for each prompt.
The dataset contains 3357 training examples and 895 evaluation examples. This dataset has a high normalised edit distance of approximately 90\%.

\paragraph{HellaSwag}
Finally, we consider the HellaSwag dataset \citep{zellers2019hellaswag},
a dataset containing commonsense inference questions known to be hard for LLMs. 
An example prompt is ``Then, the man writes over the snow covering the window of a car, and a woman wearing winter clothes smiles. then''
And the potential completions are
$[$ ", the man adds wax to the windshield and cuts it.", ", a person board a ski lift, while two men supporting the head of the person wearing winter clothes snow as the we girls sled.", ", the man puts on a christmas coat, knitted with netting.", ", the man continues removing the snow on his car." $]$
The dataset contains 119\,715 training and 30\,126 evaluation examples.

\section{Additional Experiments and Details} \label{app:experiments}

\subsection{Additional Training Details}

No hyperparameter tuning was done when creating \model-34B or \model-72B.
\methodShort{} has two hyperparameters, $\beta$ and $\lambda$.
We chose $\beta=0.3$, similar to prior work \citep{rafailov2023direct},
and we chose $\lambda=50$ without trying other values.
It is possible that even better performance can be achieved, e.g., with a different value of $\lambda$.

Here, we give the licenses of all models used to train our \model-series of models.

\model-7B started from Llama 2-chat \citep{touvron2023llama2}.
Therefore, we release it under the Llama 2 license (\url{https://ai.meta.com/llama/license/}).

\model-34B started from Bagel-34B-v0.2 \citep{bagel34B}, which itself is a SFT version of Yi-34B-200k \citep{Yi34B200K}.
Therefore, we release \model-34B under the Yi Series Models Community License Agreement (\url{https://github.com/01-ai/Yi/blob/main/MODEL_LICENSE_AGREEMENT.txt}).

\model-72B started from MoMo-72b-lora-1.8.7-DPO \citep{MoMo72B}, which itself is a fine-tune of Qwen-72B \citep{bai2023qwen}.
Therefore, we release \model-72B under the Qwen license (\url{https://github.com/QwenLM/Qwen/blob/main/Tongyi%20Qianwen%20LICENSE%20AGREEMENT}).

\subsection{Additional Results}

\begin{figure}[t]
    \centering
    \includegraphics[width=.5\linewidth]{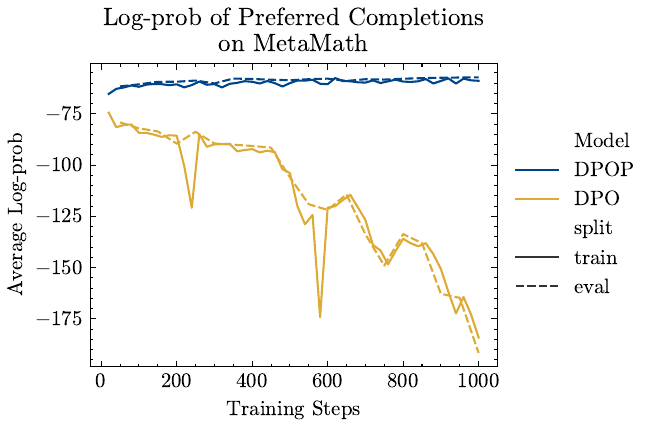}
    \caption{Average log-probs for preferred completions of \methodShort{} and DPO on the MetaMath dataset showing the failure mode of DPO.
    }
    \label{fig:dpo_failures}
\end{figure}

\begin{figure}[t]
    \centering
    \includegraphics[width=.5\linewidth]{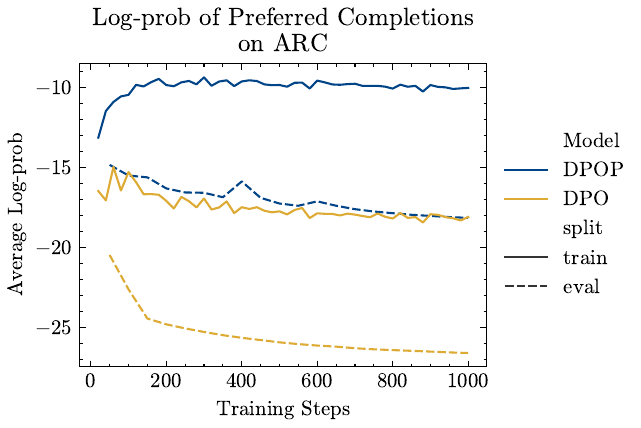}
    \caption{Average log-probs for preferred completions of \methodShort{} and DPO on the ARC dataset showing the failure mode of DPO.
    }
    \label{fig:dpo_failures_arc}
\end{figure}

\paragraph{Log-probabilities of preferred completions} In \cref{fig:dpo_failures}, we show the log-probabilities of the preferred completion of the train and eval sets during training on MetaMath. 
We plot the log-probabilities in more granularity than in \cref{fig:logprob_location}.
We confirm our theoretical insights from \cref{sec:dpop} -- the log-probabilities of the preferred completion drop substantially in DPO, whereas they increase for \methodShort{} -- across both the train and eval sets. For ARC, we see in \cref{fig:dpo_failures_arc} that \methodShort{} maintains high train-set log-probs, while both the train and eval set log-probs decrease for DPO. Notably, even though eval set log-probs do decrease for \methodShort{}, they are still higher than the train set log-probs of DPO.

\paragraph{Additional tables}
In \cref{tab:leaderboard_extension}, we give an extension of \cref{tab:leaderboard} (whose experimental details are in \cref{sec:smaug}).
In \cref{tab:leaderboard_34b}, we give the same table, except for models of size 34B or lower.

\paragraph{MT-Bench}
Next, we evaluate using MT-Bench \citep{zheng2023judging}, a challenging benchmark that uses GPT-4 \citep{gpt4} to score candidate model responses across eight different categories of performance. 
As shown in other works \citep{zheng2023judging, rafailov2023direct}, strong LLMs such as GPT-4 show good agreement with human preferences. 
We note that MT-Bench questions were not included in our fine-tuning datasets in any way.
We run MT-Bench with the Llama-2 conversation template \citep{touvron2023llama2}. 
%We averaged multi-turn
See Appendix \cref{tab:mtbench} for a comparison with state-of-the-art LLMs according to Arena Elo as of March 2024. 
\model-72B achieves the top MMLU score and third-best MT-bench score out of the open-source models. 
In \cref{app:experiments}, we give examples of \model-72B completions to MT-Bench questions.

\begin{table*}[t]
\caption{Evaluation of the top models according to MT-Bench and MMLU.}
\label{tab:mtbench}
\resizebox{\textwidth}{!}{%
\centering
\begin{tabular}{lrrrr}
\toprule
Model & MT-bench & MMLU & Organization & License \\
\midrule
GPT-4-1106-preview & \textbf{9.32} & & OpenAI & Proprietary \\
GPT-4-0314 & 8.96 & \textbf{86.4} & OpenAI & Proprietary \\
GPT-4-0613 & 9.18 & & OpenAI & Proprietary \\
Mistral Medium & 8.61 & 75.3 & Mistral & Proprietary \\
Claude-1 & 7.9 & 77 & Anthropic & Proprietary \\
Claude-2.0 & 8.06 & 78.5 & Anthropic & Proprietary \\
Gemini Pro (Dev API) & & 71.8 & Google & Proprietary \\
Claude-2.1 & 8.18 & & Anthropic & Proprietary \\
GPT-3.5-Turbo-0613 & 8.39 & & OpenAI & Proprietary \\
Mixtral-8x7b-Instruct-v0.1 & 8.3 & 70.6 & Mistral & Apache 2.0 \\
Yi-34B-Chat & & 73.5 & 01 AI & Yi License \\
Gemini Pro & & 71.8 & Google & Proprietary \\
Claude-Instant-1 & 7.85 & 73.4 & Anthropic & Proprietary \\
GPT-3.5-Turbo-0314 & 7.94 & 70 & OpenAI & Proprietary \\
WizardLM-70B-v1.0 & 7.71 & 63.7 & Microsoft & Llama 2 Community \\
Tulu-2-DPO-70B & 7.89 & & AllenAI/UW & AI2 ImpACT Low-risk \\
Vicuna-33B & 7.12 & 59.2 & LMSYS & Non-commercial \\
Starling-LM-7B-alpha & 8.09 & 63.9 & UC Berkeley & CC-BY-NC-4.0 \\
\midrule
\textbf{\model-72B} & 7.76 & 77.15 & Abacus.AI & tongyi-qianwen-license-agreement \\
\bottomrule
\end{tabular}
}
\end{table*}

\subsection{Contamination Checker} \label{app:contamination}
Since LLMs train on wide swaths of the internet, data contamination, i.e., evaluating on examples that are included in the training data, is a growing concern.
Data contamination remains notoriously challenging to measure and mitigate, even with partial attempts like controlled experimentation of training data, canary strings, or embedding similarities \citep{roberts2024cutoff,jain2024livecodebench,srivastava2023imitation}.

While there is no perfect tool to check for data contamination, we use an open-source contamination checker (\url{https://github.com/swj0419/detect-pretrain-code-contamination}) to compare the contamination of our model to other open-source models - see \cref{tab:contamination}. On ARC, TruthfulQA, and GSM8K, we find that \model-72B achieves scores that are similar to MoMo-72B-lora-1.8.7-DPO (the starting point of \model-72B), as well as Llama-2-70B.

\begin{table}[t]
\centering
\begin{tabular}{|l|c|c|c|}
\hline
\textbf{Model} & \textbf{\model-72B} & \textbf{MoMo-72B-lora-1.8.7-DPO} & \textbf{Llama-2-70B} \\
\hline
ARC & 0.20 & 0.20 & 0.22 \\
TruthfulQA & 0.45 & 0.39 & 0.51 \\
GSM8K & 1.00 & 1.00 & 0.89 \\
\hline
\end{tabular}
\caption{Comparison of contamination values across models, according to
\url{https://github.com/swj0419/detect-pretrain-code-contamination}. A score above 0.85 indicates likely contamination.
}
\label{tab:contamination}
\end{table}

\begin{table*}[t]
\caption{(Extension of \cref{tab:leaderboard}.) Evaluation of the top open-weight models on the HuggingFace Open LLM Leaderboard as of March 2024.
Our 72B model achieves an average accuracy of 80.48\%, becoming the first open-source LLM to surpass an average accuracy of 80\% and improving by nearly 2\% over the second-best open-source model (other than a fine-tune of our own model).
}
\label{tab:leaderboard_extension}
\resizebox{\textwidth}{!}{%
\centering
\begin{tabular}{lrrrrrrr}
\toprule
Model & Avg.\ & ARC & HellaSwag & MMLU & TruthfulQA & Winogrande & GSM8K \\
\midrule
\textbf{\model-72B (Ours)} & \textbf{80.48} & \textbf{76.02} & 89.27 & 77.15 & \textbf{76.67} & 85.08 & \textbf{78.70} \\
MoMo-72B-lora-1.8.7-DPO & 78.55 & 70.82 & 85.96 & 77.13 & 74.71 & 84.06 & 78.62 \\
TomGrc\_FusionNet\_34Bx2\_MoE\_v0.1\_DPO\_f16 & 77.91 & 74.06 & 86.74 & 76.65 & 72.24 & 83.35 & 74.45 \\
TomGrc\_FusionNet\_34Bx2\_MoE\_v0.1\_full\_linear\_DPO & 77.52 & 74.06 & 86.67 & 76.69 & 71.32 & 83.43 & 72.93 \\
Truthful\_DPO\_TomGrc\_FusionNet\_7Bx2\_MoE\_13B & 77.44 & 74.91 & \textbf{89.30} & 64.67 & 78.02 & \textbf{88.24} & 69.52 \\
CCK\_Asura\_v1 & 77.43 & 73.89 & 89.07 & 75.44 & 71.75 & 86.35 & 68.08 \\
FusionNet\_34Bx2\_MoE\_v0.1 & 77.38 & 73.72 & 86.46 & 76.72 & 71.01 & 83.35 & 73.01 \\
MoMo-72B-lora-1.8.6-DPO & 77.29 & 70.14 & 86.03 & \textbf{77.40} & 69.00 & 84.37 & 76.80 \\
\textbf{\model-34B-v0.1 (Ours)} & 77.29 & 74.23 & 86.76 & 76.66 & 70.22 & 83.66 & 72.18 \\
Truthful\_DPO\_TomGrc\_FusionNet\_34Bx2\_MoE & 77.28 & 72.87 & 86.52 & 76.96 & 73.28 & 83.19 & 70.89 \\
DARE\_TIES\_13B & 77.10 & 74.32 & 89.5 & 64.47 & 78.66 & 88.08 & 67.55 \\
13B\_MATH\_DPO & 77.08 & 74.66 & 89.51 & 64.53 & 78.63 & 88.08 & 67.10 \\
FusionNet\_34Bx2\_MoE & 77.07 & 72.95 & 86.22 & 77.05 & 71.31 & 83.98 & 70.89 \\
MoE\_13B\_DPO & 77.05 & 74.32 & 89.39 & 64.48 & 78.47 & 88.00 & 67.63 \\
4bit\_quant\_TomGrc\_FusionNet\_34Bx2\_MoE\_v0.1\_DPO & 76.95 & 73.21 & 86.11 & 75.44 & 72.78 & 82.95 & 71.19 \\
MixTAO-7Bx2-MoE-Instruct-v7.0 & 76.55 & 74.23 & 89.37 & 64.54 & 74.26 & 87.77 & 69.14 \\
Truthful\_DPO\_cloudyu\_Mixtral\_34Bx2\_MoE\_60B0 & 76.48 & 71.25 & 85.24 & 77.28 & 66.74 & 84.29 & 74.07 \\
MoMo-72B-lora-1.8.4-DPO & 76.23 & 69.62 & 85.35 & 77.33 & 64.64 & 84.14 & 76.27 \\
FusionNet\_7Bx2\_MoE\_v0.1 & 76.16 & 74.06 & 88.90 & 65.00 & 71.20 & 87.53 & 70.28 \\
MBX-7B-v3-DPO & 76.13 & 73.55 & 89.11 & 64.91 & 74.00 & 85.56 & 69.67 \\
\bottomrule
\end{tabular}
}
\end{table*}

\begin{table*}[t]
\caption{Evaluation of the top open-weight models on the HuggingFace Open LLM Leaderboard for models less than 35B parameters as of March 2024. Our 34B model achieves best-in-its-class performance compared to other models of similar or smaller size.
}
\label{tab:leaderboard_34b}
\resizebox{\textwidth}{!}{%
\centering
\begin{tabular}{lrrrrrrrr}
\toprule
Model & Size &  Avg.\ & ARC & HellaSwag & MMLU & TruthfulQA & Winogrande & GSM8K \\
\midrule
\textbf{\model-34B-v0.1 (Ours)} & <35B & \textbf{77.29} & 74.23 & 86.76 & \textbf{76.66} & 70.22 & 83.66 & \textbf{72.18} \\
DARE\_TIES\_13B & <35B & 77.10 & 74.32 & 89.50 & 64.47 & \textbf{78.66} & \textbf{88.08} & 67.55 \\
13B\_MATH\_DPO & <35B & 77.08 & \textbf{74.66} & \textbf{89.51} & 64.53 & 78.63 & \textbf{88.08} & 67.10 \\
MoE\_13B\_DPO & <35B & 77.05 & 74.32 & 89.39 & 64.48 & 78.47 & 88.00 & 67.63 \\
4bit\_quant\_TomGrc\_FusionNet\_34Bx2\_MoE\_v0.1\_DPO & <35B & 76.95 & 73.21 & 86.11 & 75.44 & 72.78 & 82.95 & 71.19 \\
\bottomrule
\end{tabular}
}
\end{table*}

\section{Example Completions} \label{app:examples}
In this section, we give example completions by \model-72B for questions in MT-Bench \citep{zheng2023judging}.
Note that these are not cherry-picked -- they include examples of both good and bad completions.

\onecolumn

\begin{tcolorbox}[colback=white, colframe=darkgray, title={\textcolor{white}{Example 1}}, fonttitle=\bfseries]
\vspace{-0.9em}
\tcbsubtitle{{\textcolor{white}{Question: }\textcolor{aquamarine}{Imagine you are writing a blog post comparing two popular smartphone models. Develop an outline for the blog post, including key points and subheadings to effectively compare and contrast the features, performance, and user experience of the two models. Please answer in fewer than 200 words.}}}

Outline:
\begin{enumerate}[topsep=2pt, itemsep=2pt, parsep=0pt, leftmargin=5mm]
\item Introduction to the two smartphone models being compared (mention brand names and models)
\item Design and Build Quality
\begin{itemize}[topsep=2pt, itemsep=2pt, parsep=0pt, leftmargin=5mm]
\item Comparison of dimensions, weight, materials used, and overall aesthetic appeal
\item Water and dust resistance, if applicable
\end{itemize}
\item Display and Screen Quality
\begin{itemize}[topsep=2pt, itemsep=2pt, parsep=0pt, leftmargin=5mm]
\item Comparison of screen technology (OLED, IPS LCD, etc.)
\item Display resolution, refresh rate, and brightness levels
\end{itemize}
\item Performance and Hardware
\begin{itemize}[topsep=2pt, itemsep=2pt, parsep=0pt, leftmargin=5mm]
\item Processor, RAM, and storage configurations
\item Comparison of benchmark scores (e.g., AnTuTu, Geekbench)
\end{itemize}
\item Camera System and Image Quality
\begin{itemize}[topsep=2pt, itemsep=2pt, parsep=0pt, leftmargin=5mm]
\item Comparison of primary, ultra-wide, telephoto, and macro lenses (if available)
\item Comparison of low-light performance and image processing capabilities
\end{itemize}
\item Battery Life and Charging
\begin{itemize}[topsep=2pt, itemsep=2pt, parsep=0pt, leftmargin=5mm]
\item Battery capacity and estimated usage time
\item Charging technologies (wired, wireless, fast charging, reverse charging)
\end{itemize}
\item Software and User Experience
\begin{itemize}[topsep=2pt, itemsep=2pt, parsep=0pt, leftmargin=5mm]
\item Comparison of operating systems (iOS vs. Android) and their features
\item User interface, customization options, and app ecosystem
\end{itemize}
\item Additional Features and Extras
\begin{itemize}[topsep=2pt, itemsep=2pt, parsep=0pt, leftmargin=5mm]
\item Comparison of biometric authentication (fingerprint, face unlock, etc.)
\item Additional features like NFC, 5G connectivity, and audio quality
\end{itemize}
\item Pricing and Value for Money
\begin{itemize}[topsep=2pt, itemsep=2pt, parsep=0pt, leftmargin=5mm]
\item Comparison of retail prices and current offers
\item Assessment of overall value considering the features and performance
\end{itemize}
\item Conclusion and recommendation for potential buyers based on their preferences
\end{enumerate}

\end{tcolorbox}

\begin{tcolorbox}[colback=white, colframe=darkgray, title={\textcolor{white}{Example 2}}, fonttitle=\bfseries]
\vspace{-0.9em}
\tcbsubtitle{{\textcolor{white}{Question: }\textcolor{aquamarine}{In a survey conducted at a local high school, preferences for a new school color were measured: 58\% of students liked the color blue, 45\% preferred green, and 22\% liked both colors. If we randomly pick a student from the school, what's the probability that they would like neither blue nor green?}}}

To find the probability that a randomly picked student would like neither blue nor green, we need to find the percentage of students who like only one color or neither color.

From the given information, we know:

\begin{itemize}[topsep=2pt, itemsep=2pt, parsep=0pt, leftmargin=5mm]
\item 58\% like blue
\item 45\% prefer green
\item 22\% like both colors
\end{itemize}
To find the percentage of students who like only one color, we can subtract the percentage of students who like both colors from the sum of students who like blue and green:

\begin{itemize}[topsep=2pt, itemsep=2pt, parsep=0pt, leftmargin=5mm]
\item 58\% (blue) + 45\% (green) - 22\% (both) = 83\%
\end{itemize}

Now, to find the probability of a student liking neither blue nor green, we subtract the percentage of students who like at least one color from 100\%:

\begin{itemize}[topsep=2pt, itemsep=2pt, parsep=0pt, leftmargin=5mm]
\item 100\% - 83\% = 17\%
\end{itemize}

So, the probability that a randomly picked student would like neither blue nor green is 17\%.

\end{tcolorbox}

\begin{tcolorbox}[colback=white, colframe=darkgray, title={\textcolor{white}{Example 3}}, fonttitle=\bfseries]
\vspace{-0.9em}
\tcbsubtitle{{\textcolor{white}{Question: }\textcolor{aquamarine}{Implement a program to find the common elements in two arrays without using any extra data structures.}}}

Finding common elements in two arrays without using extra data structures can be achieved by using a simple algorithm. Here's a Python implementation of this problem:

\begin{lstlisting}[language=Python]
def find_common_elements(arr1, arr2):
    common_elements = []
    for num1 in arr1:
        if num1 in arr2 and num1 not in common_elements:
            common_elements.append(num1)
    return common_elements

# Example usage:
array1 = [1, 2, 3, 4, 5]
array2 = [4, 5, 6, 7, 8]
print(find_common_elements(array1, array2))  # Output: [4, 5]
\end{lstlisting}

This program iterates through the elements of the first array (arr1) and checks if each element is present in the second array (arr2). If it is, and the element hasn't been added to the common\_elements list yet, it is appended. The function then returns the list of common elements.

\end{tcolorbox}

\begin{tcolorbox}[colback=white, colframe=darkgray, title={\textcolor{white}{Example 4}}, fonttitle=\bfseries]
\vspace{-0.9em}
\tcbsubtitle{{\textcolor{white}{Question: }\textcolor{aquamarine}{Identify the named entities (people, organizations, locations) mentioned in the given news article. Please generate a JSON dictionary that lists the named entities in three separate groups based on their entity types. The key is the type of entity and the value is a list of strings.\\
\\
Yesterday, Adamson Emerson, the CEO of Faraday, and Dieter Zetsche, the CEO of Daimler AG, announced plans to build a new Gigafactory in Berlin. The facility will be a joint venture between Faraday and Daimler, producing electric vehicles and battery packs for both companies, creating thousands of job opportunities in the region. Emerson and Zetsche stated that the strategic location of Berlin, coupled with its skilled workforce and strong infrastructure, makes it an ideal choice for expansion. The new Gigafactory aims to meet the growing demand for electric vehicles in Europe and contribute to a sustainable future. Volkswagen CEO Herbert Diess welcomed the news, saying greater collaboration will, benefit the auto industry's transition to e-mobility.}}}

\begin{lstlisting}[language=Python]
{
  "people": ["Adamson Emerson", "Dieter Zetsche", "Herbert Diess"],
  "organizations": ["Faraday", "Daimler AG", "Volkswagen"],
  "locations": ["Berlin", "Europe"]
}
\end{lstlisting}

\end{tcolorbox}

\end{document}